%% file: Batista20.tex
\documentclass[conference]{IEEEtran}
\IEEEoverridecommandlockouts
\usepackage{cite}
\usepackage{amsmath,amssymb,amsfonts}
\usepackage{algorithmic}
\usepackage{graphicx}
\usepackage{textcomp}
\usepackage{xcolor}
\usepackage{parskip}
\def\BibTeX{{\rm B\kern-.05em{\sc i\kern-.025em b}\kern-.08em
    T\kern-.1667em\lower.7ex\hbox{E}\kern-.125emX}}

\usepackage{array}
\newcolumntype{P}[1]{>{\centering\arraybackslash}p{#1}}

\definecolor{myGreen}{rgb}{0.25,0.60,0.15}
\definecolor{myYellow}{rgb}{0.85,0.65,0.0}
\definecolor{myRed}{rgb}{0.85,0.40,0.30}
\definecolor{myLightBlue}{rgb}{0.0,0.5,0.9}
\definecolor{myBlue}{rgb}{0.0,0.3,0.6}

\newcommand{\green}[1]{{\color{myGreen}#1}}
\newcommand{\yellow}[1]{{\color{myYellow}#1}}
\newcommand{\red}[1]{{\color{myRed}#1}}

\usepackage{paralist}
\usepackage[export]{adjustbox}

\usepackage{amsfonts} %% <- also included by amssymb
\DeclareMathSymbol{\shortminus}{\mathbin}{AMSa}{"39}

\begin{document}

\title{
    Improving the Detection of Burnt Areas in Remote Sensing using Hyper-features Evolved by M3GP
    \thanks{
        Work funded by Funda\c{c}\~{a}o para a Ci\^{e}ncia e a Tecnologia (FCT), Portugal.
    }
}

\author{
\IEEEauthorblockN{Jo\~{a}o E. Batista, Sara Silva}
\IEEEauthorblockA{\textit{LASIGE, Faculdade de Ci\^{e}ncias, Universidade de Lisboa, Portugal}
\\
\{jebatista,sara\}@fc.ul.pt}
}

\maketitle

\input{Tex/A_abstract_keywords.tex}
\input{Tex/B_introduction.tex}
\input{Tex/C_relatedWork.tex}
\input{Tex/E_m3gp.tex}

\input{Tex/G_experimentalSetup.tex}
\input{Tex/I_results.tex}

\input{Tex/K_discussion.tex}
\input{Tex/L_conclusions_futurework.tex}
\input{Tex/O_acknowledgment.tex}

\bibliography{Batista20}{}
\bibliographystyle{IEEEtran}

\end{document}

%% file: Tex/A_abstract_keywords.tex
\begin{abstract}
One problem found when working with satellite images is the radiometric variations across the image and different images.
Intending to improve remote sensing models for the classification of burnt areas, we set two objectives. The first is to understand the relationship between feature spaces and the predictive ability of the models, allowing us to explain the differences between learning and generalization when training and testing in different datasets. We find that training on datasets built from more than one image provides models that generalize better. These results are explained by visualizing the dispersion of values on the feature space. The second objective is to evolve hyper-features that improve the performance of different classifiers on a variety of test sets. We find the hyper-features to be beneficial, and obtain the best models with XGBoost, even if the hyper-features are optimized for a different method.  %Já voltei a por como estava

\end{abstract}


%% file: Tex/B_introduction.tex
\section{Introduction}
\vspace{-3pt}
Deforestation has serious implications on biodiversity, on rural communities that depend on forests for survival, and on greenhouse gas emissions that drive the global climate. One of the proposed solutions to prevent deforestation is the use of financing mechanisms, like the REDD+\footnote{www.un-redd.org} and the Zero Deforestation Program\footnote{www.tft-earth.org}. These mechanisms require the ability to perform forest monitoring, normally based on remote sensing (RS) data like satellite imagery. The machine learning (ML) community can help by providing predictive models that, after learning from a small sample of an image, can automatically classify the whole image.
\par
Although previous ML work in forest monitoring has shown good results, the predictive models are often applied on the same location where they were learnt, i.e., the models are trained and tested in samples from the same dataset (e.g.,~\cite{massachussets_houseClassification}) or time series from the same area (e.g.,~\cite{spain_cropclassification}). In order to improve the quality of the models, they need to have a better generalization ability and obtain good results, even when applied to images from locations where they were not trained.
\par
This task can be particularly hard since each band of the satellite sensors, which translates as each feature of the reference dataset, contains noise that is originated by the difference in the amount of water in the air, ground, or vegetation; by the presence of shadows; or by the angle of incidence of the solar radiation~\cite{ndvi}. When comparing different images, these radiometric variations are intensified by differences in weather conditions, the time of the day, and the plants' growth stage.
\par 
Another challenge for the ML community generating interpretable and meaningful hyper-features, similar to indices, that facilitate the learning of ML models while also allowing the RS experts to easily understand the models. Indices are something the RS community has been using for a long time, the most popular being the Normalized Difference Vegetation Index (NDVI)~\cite{ndvi}, that measures the ``greenness'' of live vegetation. The NDVI is a ratio of wavelengths that is higher in the solar radiation reflected by surfaces that contain chlorophyll and protochlorophyll. Ratios always provide interpretable values when used in cloud-free images, which allows RS experts to study different scenes using the same index, including time series~\cite{ndviTimeSeries}. They are also useful to the ML community, as they help the learning algorithms to model the data~\cite{spain_cropclassification}.

We have the goal of improving the detection of burnt areas in satellite imagery by providing robust and interpretable classification models. For the current work, we have established two objectives:
\begin{itemize}
   \item[(1)] To understand the relationship between feature spaces and the predictive ability of the models;
   \item[(2)] To obtain hyper-features that improve the performance of ML methods in detecting burnt areas.
\end{itemize}
\par
For the first objective, we study the effects of training a model on data from multiple images with different radiometric characteristics, rather than from a single image. For this task, we use Genetic Programming (GP)~\cite{Po08}, more specifically the M3GP method~\cite{m3gp}. By design, the performance of M3GP is not affected by the lack of data normalization (which could alter the radiometric biases of the images). We visualize some characteristics of the feature space created by each image dataset and relate it with the observed learning and generalization ability of each evolved model, intuitively explaining the relationship between the dispersion of the feature values and the accuracy of the predictions.
\par 
For the second objective, we evolve hyper-features from the original ones and compare the performance of different ML methods when using the evolved versus the original features. For this task, we also use M3GP because it naturally evolves hyper-features during its learning process. The ML methods tested include Decision Trees, Random Forests and XGBoost.

%% file: Tex/C_relatedWork.tex
\section{Related Work}
\vspace{-3pt}
The ML literature is filled with implicit studies of feature spaces. Nearly every classification method is based on how to best split the feature space into regions that correspond to classes. A large body of work dedicated to feature and instance selection and construction~\cite{10.5555/551943,10.5555/558166} also implicitly studies feature spaces, as well as work dedicated to data visualization techniques (review in~\cite{7784854}). In remote sensing, we have found one explicit study of feature spaces in the context of classification~\cite{10.1117/12.196749}. Published in 1994, in this work the authors visualize the feature space and the decision boundaries induced by neural and statistical classifiers, with the goal of understanding why a classifier performs better than another on a given problem. Image sequences are used to visualize higher-dimensional feature spaces.
\par
Regarding the generation of hyper-features, it falls on a broad area called Constructive Induction (CI)~\cite{Wnek1994}. A recent publication on transfer learning~\cite{transfer} has surveyed GP-based CI methods, finding a considerable amount of old and recent work where all the authors propose similar methods, apparently unaware of the existence of the other related works. Among the CI methods surveyed are the EFS~\cite{efs}, the FFX~\cite{ffx}, and the M3GP~\cite{m3gp}. M3GP is the one chosen by the authors for generating the hyper-features for transfer learning, and also the one we use in the current work.
\par
In the RS domain, many techniques have been used to extract features from imagery. Just to name a few, old and recent, the gray level co-occurrence matrix (GLCM) and other methods~\cite{GONG1992137} have been used to extract statistical descriptors from images. Many methods for extracting features of interest (e.g., roads) have been used in the past~\cite{review_articlee}, now almost completely replaced by deep learning techniques (e.g.,~\cite{fe_with_dl}). PCA is still very common in the RS domain to extract a set of generic features from the original ones (e.g.,~\cite{PCA_RS}). Temporal features prove to be very helpful in tasks of forest classification~\cite{PASQUARELLA2018193}.

%% file: Tex/E_m3gp.tex
\section{M3GP Algorithm}
\vspace{-3pt}
\label{m3gp}

M3GP~\footnote{Simple M3GP implementation: github.com/jespb/Python-M3GP}~\cite{m3gp} is a cluster-based algorithm that evolves models similar in structure to the models of standard GP~\cite{Po08}. The main difference is that, while standard GP models are limited to one node at their root, making their output 1-dimensional, the models of M3GP are allowed to have as many nodes at their root as the evolution chooses. In previous M3GP work, these nodes (including their respective branches) were normally called dimensions, but here we will also call them hyper-features. A M3GP model uses its evolved hyper-features to convert the original feature space into a new hyper-feature space. We will call hyper-dataset to the dataset made of these transformed values. Depending on the evolutionary process, the number of hyper-features may decrease, increase or remain the same, when compared to the original ones.
\par
The genetic operators of M3GP include the standard subtree crossover and mutation, as well as one additional crossover that swaps entire dimensions between individuals, and two additional mutations that add/remove a dimension to/from an individual. The fitness of a M3GP model is obtained as follows: clusters are formed on the hyper-feature space, one per class; the centroids of the clusters are calculated; the predicted class of each observation is the class of the nearest centroid, according to the Mahalanobis distance; the fitness is the accuracy of this predicted classification.
The Mahalanobis distance is preferred because it returns much better results than the Euclidean distance~\cite{m2gp}. In order to measure the fitness of a M3GP model on unseen data, one needs to know not only the structure of the $n$-rooted tree (the $n$ evolved hyper-features), but also the cluster centroids, and the covariance matrices needed to calculate the Mahalanobis distance. It may not be the most practical model to use, but M3GP has shown competitive results with other highly ranked ML methods, such as Multilayer Perceptron and Random Forests~\cite{m3gp, rodrigues20}, also in multiclass classification problems~\cite{m3gp,mgp_multiclass}.
\par
However, M3GP does not have to depend on cluster centroids or covariance matrices. By evolving hyper-features that cause the class clusters to be more easily separable, it is expected that the resulting hyper-dataset is easier to learn than the original dataset, so the transformation defined by the evolved hyper-features is by itself a very useful output of this method. Furthermore, in~\cite{m3gp_reg, transfer} M3GP was successfully used as a wrapper method for other regression and classification methods. Indeed, the fitness function of M3GP, that now uses the Mahalanobis distance nearest centroid classifier (which we will now call MD classifier) to obtain predictions, can be replaced by any other classifier. Since the hyper-features are evolved specifically for that classifier, they are expected to be optimized for that classifier.

At the end of each M3GP run, we apply two pruning procedures to the best individual to first decrease the dimensionality of the model without decreasing its fitness, and then simplify each of the remaining dimensions. While the first operation was included in the original M3GP algorithm, the second was added in this work:\\
(1) Dimensional Pruning: One by one, each dimension is temporarily removed and the individual re-evaluated. If the fitness is not harmed, the dimension is permanently removed.\\
(2) Dimensional Simplification: Any expressions that are evaluated as constants are replaced by those constants. The following rules are applied to any expression $E$:\\
{$E+0=E, E+E=2*E, E-0=E, 1*E=E, E/1=E$}.\\

%% file: Tex/G_experimentalSetup.tex
\vspace{-15pt}
\section{Experimental Setup}
\vspace{-5pt}
\subsection{Datasets}
We use three binary classification datasets, collected from Landsat-8 images and labelled by experts in the context of previous work~\cite{CABRAL201894}. The images were captured over Brazil(B), Congo(C), and Mozambique(M). The 20 pixels found mislabeled in the Brazil dataset~\cite{B} were corrected. Table~\ref{datasets} summarizes the main characteristics of these ``pure'' datasets.
\par
In order to study the effect of training a model on data from multiple images, we also use mixed datasets containing pixels from two or three pure datasets. From now on, the pure datasets are identified by their initials (B,C,M) and the mixed datasets by a concatenation of initials, e.g., a model trained on a dataset containing samples from Brazil and Congo is referred to as a model trained in BC. We may refer to models trained on a single pure dataset as specialized models.
\par
Each model is trained with 2000 samples. When these samples come from different images, each of the pure datasets contributes a number of samples that is proportional to its size, e.g., when selecting samples for the BM dataset, B contributes with 1113 samples and M with 887 samples. The same type of stratification is applied when selecting samples from the two classes. The test set is always composed of all the remaining samples. This means that, in most cases, the test set is substantially larger than the training set. A new training/test partition is made for each run.

\begin{table}
\centering
\caption{Features, Samples and \% Burnt}
\resizebox{0.98\linewidth}{!}{
\begin{tabular}{p{1.5cm}|P{1.25cm}P{1.25cm}P{1.9cm}}
Datasets      & Brazil(B)    & Congo(C)     & Mozambique(M)      \\ 
\hline
Features       & 7           & 7         & 7               \\
Samples        & 4872        & 2849      & 3882            \\
\% Burnt   & 42\%       & 31\%     & 41\%           \\
\end{tabular}
}
\label{datasets}
\vspace{-5pt}
\end{table}

\subsection{Parameters}
\vspace{-5pt}
The parameter settings are specified in Table~\ref{params}. We use  smaller populations and fewer generations than the original M3GP~\cite{m3gp} because the preliminary results indicated that more than 200 individuals and 50 generations would not improve the results. The fitness function is the accuracy, but the smaller individuals are preferred whenever there is a tie. The probability of choosing crossover or mutation is 50\% for each, and then equal inside each type, resulting in the 25\% and 16.7\% indicated in the table. All other parameters are standard, except the pruning options described at the end of Sect.~\ref{m3gp}.

\begin{table}[]
\centering
  \caption{Parameters used by M3GP}
  \resizebox{0.98\linewidth}{!}{
  \small{
  \begin{tabular}{ll}
    \hline                                        
    Runs                        & 30\\
    Generations                 & 50\\
    Population Size             & 200 individuals\\
    Training Set Size           & 2000 samples\\
    Tree Initialization         & Grow method (max depth 6)\\
    Function Set                & +, $\shortminus$, $\times$, // (protected)\\
    Terminal Set                & Dataset features (no constants) \\
    Fitness                     & Accuracy (untied by smaller size)\\
    Selection                   & Tournament (size 5)\\
    Elitism                     & Best individual\\
    Crossover Probability       & 25\% for each of the 2 crossovers\\
    Mutation Probability        & 16.7\% for each of the 3 mutations\\
    \hline
  \end{tabular}
  }
  }
  \label{params}
\vspace{-11pt}
\end{table}

\subsection{Hyper-datasets}
\vspace{-5pt}
\label{hyper}
\par
To compare the performance of different ML methods when using the evolved versus the original features, 
we built new hyper-datasets that resulted from applying the hyper-features in Table~\ref{hyper} to each of the original (pure and mixed) datasets. These hyper-features were picked from the models obtained in the 30 runs on the BCM mixed dataset in the following way: each model is evaluated with all its dimensions; for each dimension, the model is evaluated with all but this dimension and the reduction in training accuracy is registered; all the dimensions from the 30 models are ranked by their impact on the accuracy, and the 10 most impactful are chosen as hyper-features to build the hyper-datasets. While most dimensions had an impact of less than 5 percentage points on the accuracy, some had an impact of more than 20 percentage points. 

\begin{table}
\centering
\caption{Simplified hyper-features with names and originating runs}
\resizebox{\linewidth}{!}{
\begin{tabular}{@{}c@{}|@{}c@{}|@{}l@{}}
 Run~    & ~Name~            & ~Evolved hyper-feature      \\ 
\hline

1   & HF0 &~ $ {\scriptstyle X6^2 \cdot (X0 \shortminus X3) / X4^2 } $  \\
3   & HF1 &~ $ {\scriptstyle (X3 \shortminus (X3 \shortminus X4^2 \cdot (X3 \shortminus 2 \cdot X5) \cdot ( X5^2 \shortminus X3 \cdot X5 + X4 ))/X0 \shortminus X4^2 \cdot X5^2)/X5 } $   \\
7   & HF2 &~ $ {\scriptstyle X2 \cdot (X0 \cdot X2 + X1 \cdot X2^2 \cdot X6 \shortminus X2^2 \shortminus X0 \cdot X1 \cdot X2 \cdot X6 + 1 ) }$  \\
7   & HF3 &~ $ {\scriptstyle X3 \shortminus X0 \cdot X1 \shortminus X5 \cdot X6 } $          \\
12  & HF4 &~ $ {\scriptstyle ( ( X6^4 / ( X1 \shortminus X6 + X5 ) ) \shortminus X0 * X6^3 ) * X4^{11} / X0^{15} * ( X5 \shortminus X3 )^2 } $                                  \\
12  & HF5 &~ $ {\scriptstyle X4/X0 + X0 \cdot X6 + X3 \shortminus X5 } $                \\
16  & HF6 &~ $ {\scriptstyle X6 \cdot ( X1 \shortminus X0 ) } $                     \\
26  & HF7 &~ $ {\scriptstyle X5 \cdot X6 \cdot (X3 + X5) / X4 } $   \\
26  & HF8 &~ $ {\scriptstyle X0 \shortminus X6 } $                              \\
27  & HF9 &~ $ {\scriptstyle (X4 \shortminus X2) \cdot (X5 + X6 \shortminus X0) } $   \\

\end{tabular}
}
\label{hyper}
\vspace{-15pt}
\end{table}

\subsection{Additional Methods and Tools}
\vspace{-5pt}
Besides the M3GP method as described in Section~\ref{m3gp}, other methods are involved in our experiments:

\subsubsection{Mahalanobis Distance Classifier (MD)}
MD is the nearest centroid classifier used as fitness in M3GP and explained in Section~\ref{m3gp}. The hyper-features evolved by M3GP are therefore optimized for MD, so we have used this classifier also on the original features in order to assess the improvements brought by the evolved features. 

\subsubsection{Decision Trees (DT), Random Forests (RF) \& XGBoost (XGB)}
In order to assess the improvements brought by the evolved features to other ML methods for which the hyper-features were not optimized, we have selected three well known state-of-the-art methods: DT~\cite{decisionTrees} and RF~\cite{randomForests}, with implementations from the sklearn python library~\cite{scikit-learn}, and XGB~\cite{xgb} with implementation from the xgboost python library. All were used with default parameters, except the maximum tree depth of RF that we have set as 6, the same as the initial maximum tree depth for M3GP. Unlike MD, none of these three classifiers is distance-based.  

For statistical significance (Sect.~\ref{results}) we use the Kruskal-Wallis test and consider the results to be significantly different if their \textit{p}-value is lower than $0.01$. Figures~\ref{M_dist} and~\ref{BCM_burnt_dist} (Sect.~\ref{feature_spaces}) do not include outliers, identified with the Tukey's fences method.

%% file: Tex/I_results.tex
\section{Results}
\vspace{-3pt}
\label{results}
Here, we observe the generalization ability of the models on the same images where they were trained, and then, on images not used in training. We also compare the results obtained by MD, DT, RF and XGB on the original versus hyper-datasets.  
\subsection{Generalization Inside the Training Images}
Table~\ref{treinoteste} displays the median (30 runs) training and test accuracy obtained on the original (pure and mixed) datasets with the five considered methods (M3GP, MD, DT, RF, XGB). The models do not seem to overfit their training set in any of the cases. The largest difference between training and test was observed when training DT on the BCM dataset (training accuracy 100\% and test accuracy 97.37\%). In general, the accuracy is lower on the mixed datasets (BC, BM, CM, BCM), and this effect tends to be stronger when the mix includes the three countries (BCM). More specifically, in most cases, the models of the mixed datasets have lower test accuracy than the models specialized in the respective pure datasets (e.g., for M3GP the BC test accuracy is 96.25\% while the B and C test accuracies are 98.88\% and 98.93\%). Although the focus of the work is not to compare the performance of different methods, it is clear that MD is the worst while XGB seems to be the best.

Table~\ref{teste} displays the test accuracy obtained in the pure datasets (B, C, M) by models trained in all different datasets, pure and mixed. 
For now, ignore all the ``Hyper'' subtables, the colours, and the cases in which the model is tested on a dataset that was not included in the training.
We want to compare the accuracy of the models trained in pure datasets with the accuracy of the models trained with mixed datasets, when tested on the same pure datasets (e,g., when testing M3GP on B, model B achieves 98.88\% while models BC, BM, and BCM achieve 97.63\%, 97.39\%, and 94.04\%, respectively). Training on mixed datasets and testing on pure ones results in a significant decrease in accuracy in 42 out of 45 comparisons. There is also a significant improvement in two comparisons, but both with the MD method, which is not so reliable.

\input{Tabelas/accuracy_Training_and_Test.tex}

\subsection{Generalization Outside the Training Images }
\label{generalization_outside}

Here we are interested in measuring the fitness obtained on pure datasets that were never seen during training. Keep ignoring the ``Hyper'' subtables.

We want to check whether the models trained in mixed datasets achieve higher accuracy than the models trained in pure datasets. Table~\ref{teste} contains this information, where bold green means the result is significantly better than the other two cases, and italic yellow means the result is significantly better than one of the other two cases. For example, when testing RF on B, model CM achieves a significantly higher accuracy (82.0\%) than the specialized models C and M (73.46\% and 77.12\%, respectively); when testing RF on C, model BM achieves a significantly higher accuracy than model M, but model B is significantly better than models M or BM.

When tested on a new image, models trained on mixed datasets have significantly higher accuracy than the respective specialized models in 16 out of 30 cases and significantly lower accuracy in 4 out of 30 cases. This improvement is noticed particularly when testing models on the B dataset (8 out of 10 cases, against 5/10 for C and 3/10 for M).

The specialized models have significantly higher accuracy in 6/30 cases. Three of these cases are comparisons between training on B or training on BC, tested on M (introducing C in training harms the generalization in M). Other two cases are comparisons between training on B or training on BM, tested on C (introducing M in training harms the generalization on C). This apparent incompatibility between datasets C and M will be discussed later. 

Now, we take a closer look at the results restricted to the cases where the training was made on a pure dataset. We will refer to the training datasets as sources and the testing datasets as targets. On average, the accuracy of the models trained on datasets B and M is 75.90\% and 74.12\%, respectively, but only 68.88\% for the models trained on C. 
Out of the three datasets, C seems to be the worst source.
On average, the accuracy of the models tested on dataset C is 76.18\%  reduced to 72.17\% for models tested on B, and further reduced to 69.74\% when testing on M. Dataset C seems to be the best target. It is worth mentioning that all these accuracy values are worse than the ones for training on mixed datasets, which indicates that mixed datasets make better sources than pure datasets. The suitability of datasets as sources or targets will also be discussed later.

\input{Tabelas/accuracy_Test_and_Transfer.tex}

\subsection{Training and Testing on Hyper-datasets}
Now we are interested in the results obtained when using the hyper-datasets (see Sect.~\ref{hyper}), and their comparison to the results already reported for the original datasets. Table~\ref{teste} shows these results in the ``Hyper'' subtables, and Table~\ref{pvalues} contains the $p$-values of the statistical comparison, where bold green/italic red means that, for training with the method on the left and testing on the method on the top, using the hyper-features is significantly better/worse than using the original features. We can observe that using the hyper-features yields significantly better results in 17/21 cases when using MD, 13/21 when using RF, 10/21 with XGB, and 7/21 with DT. 
The effect of training with mixed hyper-datasets instead of pure hyper-datasets is even stronger than it already was for the original datasets. The accuracy is significantly better in 7/8 cases when testing on hyper-datasets B or M, and 6/8 cases when testing on hyper-dataset C, totaling 20/24 cases.

Focusing on the cases where the training was made on a pure hyper-dataset, on average the accuracy obtained on unseen hyper-datasets is 83.21\%, 69.35\%, and 76.39\% for training on hyper-datasets B, C, and M, respectively. On average, when testing on hyper-datasets B, C, and M, the accuracy is 72.43\%, 78.20\%, and 78.31\%, respectively. Once again C is a bad source but a good target. B seems like a good source but a bad target. Like with the original datasets, all these values are lower than the ones obtained when training with mixed hyper-datasets.

\input{Tabelas/pvalues_Hyper_vs_Original.tex}

%% file: Tabelas/accuracy_Training_and_Test.tex
\begin{table}[]
	\centering
	\caption{Median training and test accuracy of the five methods on each of the original (pure and mixed) datasets.}
	\resizebox{\linewidth}{!}{
	\begin{tabular}{c}

	\begin{tabular}{p{0.9cm}|P{0.6cm}P{0.6cm}P{0.6cm}P{0.6cm}P{0.6cm}P{0.6cm}P{0.8cm}|}
	M3GP&B	&C	&M	&BC	&BM	&CM	&BCM \\
	\hline 
Training&99.20\%&99.00\%&100.0\%&96.82\%&98.50\%&99.02\%&96.25\%\\
Test&98.88\%&98.93\%&99.84\%&96.25\%&98.16\%&98.73\%&95.53\%\\
\hline
	\end{tabular}
\\\\

	\begin{tabular}{p{0.9cm}|P{0.6cm}P{0.6cm}P{0.6cm}P{0.6cm}P{0.6cm}P{0.6cm}P{0.8cm}|}
	MD&B	&C	&M	&BC	&BM	&CM	&BCM \\
	\hline 
Training&96.75\%&79.33\%&92.67\%&86.05\%&89.35\%&84.58\%&84.59\%\\
Test&96.84\%&79.06\%&92.53\%&86.11\%&89.21\%&84.23\%&84.52\%\\
\hline
	\end{tabular}

\\\\

	\begin{tabular}{p{0.9cm}|P{0.6cm}P{0.6cm}P{0.6cm}P{0.6cm}P{0.6cm}P{0.6cm}P{0.8cm}|}
	DT&B	&C	&M	&BC	&BM	&CM	&BCM \\
	\hline 
Training&100.0\%&100.0\%&100.0\%&100.0\%&100.0\%&100.0\%&100.0\%\\
Test&98.83\%&99.64\%&99.89\%&97.80\%&97.91\%&99.26\%&97.37\%\\
\hline
	\end{tabular}
\\\\

	\begin{tabular}{p{0.9cm}|P{0.6cm}P{0.6cm}P{0.6cm}P{0.6cm}P{0.6cm}P{0.6cm}P{0.8cm}|}
	RF&B	&C	&M	&BC	&BM	&CM	&BCM \\
	\hline 
Training&99.17\%&99.95\%&100.0\%&98.27\%&98.92\%&99.60\%&97.77\%\\
Test&98.95\%&99.76\%&99.86\%&97.33\%&97.74\%&99.12\%&96.53\%\\
\hline
	\end{tabular}
\\\\

	\begin{tabular}{p{0.9cm}|P{0.6cm}P{0.6cm}P{0.6cm}P{0.6cm}P{0.6cm}P{0.6cm}P{0.8cm}|}
	XGB&B	&C	&M	&BC	&BM	&CM	&BCM \\
	\hline 
Training&99.90\%&99.95\%&100.0\%&99.65\%&99.65\%&100.0\%&99.02\%\\
Test&99.07\%&99.76\%&99.89\%&98.52\%&98.63\%&99.64\%&97.56\%\\
\hline
	\end{tabular}

\end{tabular}
}
\label{treinoteste}

\vspace{-11pt}
\end{table}

%% file: Tabelas/accuracy_Test_and_Transfer.tex
\begin{table*}[]
\centering
\caption{Median test accuracy from training a model in the datasets on the left and testing in the datasets above. The colors represent that the models had significantly better results \yellow{\textit{1}} or \green{\textbf{2}} times, compared to the models trying to generalize to the dataset above.}

	\resizebox{\textwidth}{!}{
\begin{tabular}{ccc}

\begin{tabular}{p{1.4cm}|ccc|}
M3GP&B&C&M\\
\hline
B&98.88\%&76.2\%&\yellow{\textit{67.5\%}}\\
C&67.86\%&98.93\%&53.43\%\\
M&\yellow{\textit{81.88\%}}&75.27\%&99.84\%\\
\hline
BC&97.63\%&94.17\%&\yellow{\textit{73.13\%}}\\
BM&97.39\%&76.18\%&99.24\%\\
CM&\yellow{\textit{82.36\%}}&98.05\%&99.24\%\\
\hline
BCM&94.04\%&93.4\%&99.09\%\\
\hline
\end{tabular}

&

\begin{tabular}{p{1.2cm}|ccc|}
MD&B&C&M\\
\hline
B&96.84\%&\yellow{\textit{67.63\%}}&57.41\%\\
C&58.33\%&79.06\%&\yellow{\textit{59.45\%}}\\
M&\green{\textbf{63.39\%}}&69.21\%&92.53\%\\
\hline
BC&92.65\%&74.91\%&\green{\textbf{90.36\%}}\\
BM&83.92\%&\green{\textbf{69.28\%}}&95.75\%\\
CM&\yellow{\textit{61.84\%}}&73.45\%&91.99\%\\
\hline
BCM&84.54\%&69.07\%&96.12\%\\
\hline
\end{tabular}

&

\begin{tabular}{p{1.4cm}|ccc|}
MD-Hyper&B&C&M\\
\hline
B&96.72\%&\green{\textbf{80.57\%}}&\yellow{\textit{92.04\%}}\\
C&58.62\%&88.85\%&59.47\%\\
M&\yellow{\textit{60.13\%}}&69.18\%&99.46\%\\
\hline
BC&94.22\%&89.24\%&\green{\textbf{97.55\%}}\\
BM&93.56\%&\yellow{\textit{72.27\%}}&99.53\%\\
CM&\green{\textbf{61.74\%}}&82.24\%&97.59\%\\
\hline
BCM&91.67\%&77.95\%&99.39\%\\
\hline
\end{tabular}

\\\\

\begin{tabular}{p{1.4cm}|ccc|}
DT&B&C&M\\
\hline
B&98.83\%&\green{\textbf{85.06\%}}&\yellow{\textit{76.33\%}}\\
C&68.0\%&99.64\%&60.78\%\\
M&77.5\%&73.35\%&99.89\%\\
\hline
BC&97.97\%&97.79\%&69.86\%\\
BM&97.52\%&\yellow{\textit{82.44\%}}&98.44\%\\
CM&\green{\textbf{81.53\%}}&99.07\%&99.54\%\\
\hline
BCM&97.07\%&96.58\%&98.13\%\\
\hline
\end{tabular}

&

\begin{tabular}{p{1.2cm}|ccc|}
RF&B&C&M\\
\hline
B&98.95\%&\green{\textbf{85.34\%}}&75.82\%\\
C&73.46\%&99.76\%&\green{\textbf{81.5\%}}\\
M&\yellow{\textit{77.12\%}}&72.85\%&99.86\%\\
\hline
BC&98.64\%&95.19\%&77.03\%\\
BM&97.21\%&\yellow{\textit{81.23\%}}&98.53\%\\
CM&\green{\textbf{82.0\%}}&99.37\%&98.93\%\\
\hline
BCM&96.04\%&96.28\%&97.58\%\\
\hline
\end{tabular}

&

\begin{tabular}{p{1.4cm}|ccc|}
XGB&B&C&M\\
\hline
B&99.07\%&\green{\textbf{83.71\%}}&\yellow{\textit{75.87\%}}\\
C&76.66\%&99.76\%&\green{\textbf{89.32\%}}\\
M&\yellow{\textit{77.5\%}}&73.13\%&99.89\%\\
\hline
BC&98.64\%&98.31\%&72.53\%\\
BM&98.32\%&\yellow{\textit{80.16\%}}&99.06\%\\
CM&\green{\textbf{80.55\%}}&99.45\%&99.74\%\\
\hline
BCM&97.65\%&95.77\%&98.64\%\\
\hline
\end{tabular}

\\\\

\begin{tabular}{p{1.4cm}|ccc|}
DT-Hyper&B&C&M\\
\hline
B&98.72\%&85.29\%&\yellow{\textit{81.1\%}}\\
C&66.7\%&99.52\%&66.17\%\\
M&\green{\textbf{85.79\%}}&\yellow{\textit{72.65\%}}&99.78\%\\
\hline
BC&97.36\%&97.39\%&\green{\textbf{87.71\%}}\\
BM&97.69\%&\yellow{\textit{86.27\%}}&99.21\%\\
CM&\yellow{\textit{83.74\%}}&99.27\%&99.7\%\\
\hline
BCM&96.67\%&96.6\%&99.14\%\\
\hline
\end{tabular}

&

\begin{tabular}{p{1.2cm}|ccc|}
RF-Hyper&B&C&M\\
\hline
B&98.97\%&79.46\%&\yellow{\textit{81.77\%}}\\
C&66.94\%&99.76\%&73.69\%\\
M&\yellow{\textit{83.69\%}}&78.78\%&99.94\%\\
\hline
BC&98.03\%&96.23\%&\green{\textbf{89.47\%}}\\
BM&98.35\%&\green{\textbf{85.24\%}}&99.28\%\\
CM&\green{\textbf{85.92\%}}&99.45\%&99.68\%\\
\hline
BCM&96.97\%&96.41\%&99.31\%\\
\hline
\end{tabular}

&

\begin{tabular}{p{1.4cm}|ccc|}
XGB-Hyper&B&C&M\\
\hline
B&98.93\%&81.41\%&84.04\%\\
C&74.97\%&99.76\%&\yellow{\textit{88.2\%}}\\
M&\yellow{\textit{82.59\%}}&\yellow{\textit{78.25\%}}&99.89\%\\
\hline
BC&98.15\%&97.37\%&\yellow{\textit{88.7\%}}\\
BM&98.33\%&\green{\textbf{86.71\%}}&99.54\%\\
CM&\green{\textbf{84.9\%}}&99.6\%&99.85\%\\
\hline
BCM&97.48\%&97.11\%&99.36\%\\
\hline
\end{tabular}

\end{tabular}

\label{teste}
}
\vspace{-10pt}
\end{table*}

%% file: Tabelas/pvalues_Hyper_vs_Original.tex
\begin{table}[htbp]
	\centering
	%\caption{P-values comparing each method when training a model on the original and hyper-datasets on the left and testing it in the dataset above. The cases where the model trained in the hyper-dataset was significantly \green{\textbf{better}} or \red{worse} are highlighted. }
	%\caption{P-values comparing each method when training a model on the original and hyper-datasets on the left and testing it in the dataset above. The cases when using the hyper-datasets lead to significantly \green{\textbf{better}} or \red{worse} are highlighted. }
	%\caption{P-values comparing the results displayed in Table~\ref{teste} when training a model on the original and hyper-datasets. The cases where using the hyper-datasets lead to significantly \green{\textbf{better}} or \red{worse} results are highlighted. }
	\caption{P-values comparing the models trained in the original and hyper-datasets. The cases where using the hyper-datasets lead to significantly \green{\textbf{better}} or \red{worse} test results are highlighted. }
	\resizebox{\linewidth}{!}{
	\begin{tabular}{cc}

\begin{tabular}{l|ccc}
MD&B	&C	&M	\\
	\hline 
B&0.237&\phantom{}\green{\textbf{0.000}}&\phantom{}\green{\textbf{0.000}}\\
C&\phantom{}\green{\textbf{0.000}}&\phantom{}\green{\textbf{0.000}}&\phantom{}\green{\textbf{0.000}}\\
M&\phantom{}\red{0.000}&\phantom{}\red{0.000}&\phantom{}\green{\textbf{0.000}}\\
\hline 
BC&\phantom{}\green{\textbf{0.000}}&\phantom{}\green{\textbf{0.000}}&\phantom{}\green{\textbf{0.000}}\\
BM&\phantom{}\green{\textbf{0.000}}&\phantom{}\green{\textbf{0.000}}&\phantom{}\green{\textbf{0.000}}\\
CM&0.315&\phantom{}\green{\textbf{0.000}}&\phantom{}\green{\textbf{0.000}}\\
\hline 
BCM&\phantom{}\green{\textbf{0.000}}&\phantom{}\green{\textbf{0.000}}&\phantom{}\green{\textbf{0.000}}\\
\hline
	\end{tabular}

&

	\begin{tabular}{l|ccc}
	DT&B	&C	&M	\\
	\hline 
B&0.050&0.359&\phantom{}\green{\textbf{0.000}}\\
C&\phantom{}\red{0.006}&0.200&0.882\\
M&\phantom{}\green{\textbf{0.000}}&0.021&0.012\\
\hline 
BC&\phantom{}\red{0.000}&\phantom{}\red{0.001}&\phantom{}\green{\textbf{0.000}}\\
BM&0.081&\phantom{}\green{\textbf{0.005}}&\phantom{}\green{\textbf{0.000}}\\
CM&\phantom{}\green{\textbf{0.000}}&0.033&0.026\\
\hline 
BCM&\phantom{}\red{0.003}&0.965&\phantom{}\green{\textbf{0.000}}\\
\hline
	\end{tabular}

\\\\

	\begin{tabular}{l|ccc}
	RF&B	&C	&M	\\
	\hline 
B&0.229&\phantom{}\red{0.000}&\phantom{}\green{\textbf{0.000}}\\
C&\phantom{}\red{0.000}&0.879&\phantom{}\red{0.003}\\
M&\phantom{}\green{\textbf{0.000}}&\phantom{}\green{\textbf{0.000}}&\phantom{}\green{\textbf{0.001}}\\
\hline 
BC&\phantom{}\red{0.000}&\phantom{}\green{\textbf{0.000}}&\phantom{}\green{\textbf{0.000}}\\
BM&\phantom{}\green{\textbf{0.000}}&\phantom{}\green{\textbf{0.000}}&\phantom{}\green{\textbf{0.000}}\\
CM&\phantom{}\green{\textbf{0.000}}&0.539&\phantom{}\green{\textbf{0.000}}\\
\hline 
BCM&\phantom{}\green{\textbf{0.000}}&0.510&\phantom{}\green{\textbf{0.000}}\\
\hline
	\end{tabular}
&

	\begin{tabular}{l|ccc}
	XGB&B	&C	&M	\\
	\hline 
B&0.017&0.198&\phantom{}\green{\textbf{0.000}}\\
C&\phantom{}\red{0.000}&0.301&\phantom{}\red{0.001}\\
M&\phantom{}\green{\textbf{0.000}}&\phantom{}\green{\textbf{0.000}}&0.129\\
\hline 
BC&\phantom{}\red{0.000}&\phantom{}\red{0.000}&\phantom{}\green{\textbf{0.000}}\\
BM&0.166&\phantom{}\green{\textbf{0.000}}&\phantom{}\green{\textbf{0.000}}\\
CM&\phantom{}\green{\textbf{0.000}}&0.072&\phantom{}\green{\textbf{0.001}}\\
\hline 
BCM&0.024&\phantom{}\green{\textbf{0.000}}&\phantom{}\green{\textbf{0.000}}\\
\hline
	\end{tabular}

\end{tabular}
}
\label{pvalues}
\end{table}

%% file: Tex/K_discussion.tex
\section{Discussion}
\vspace{-3pt}
\label{discussion}
Here, we analyze feature spaces and provide tentative explanations for some results reported in the last section, then we visualize class separability in the hyper-feature space, and finally, we address the issues of transfer learning and optimizing features for specific methods.

\subsection{Analyzing Feature Spaces of Sources and Targets}
\label{feature_spaces}
\begin{figure}[]
	\centering
	\includegraphics[width=0.94\linewidth]{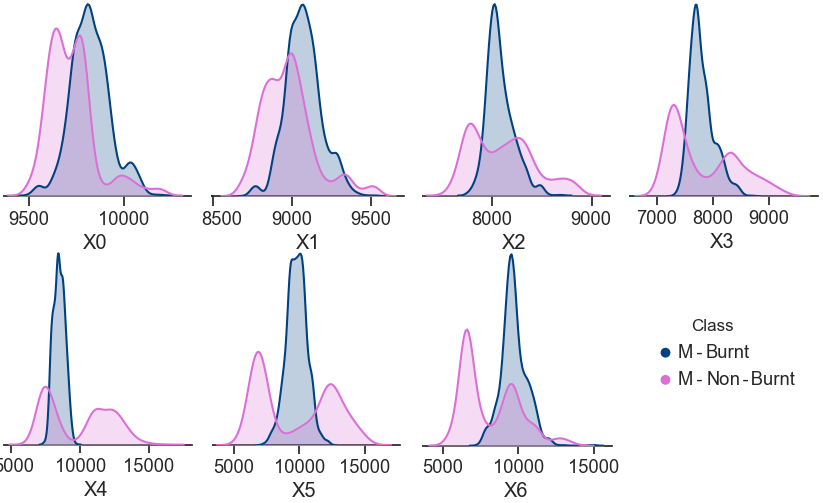}
	\caption{Distribution of feature values in the original M dataset.}
	\label{M_dist}
\vspace{-12pt}
\end{figure}

\begin{figure}[]
	\centering

	\includegraphics[width=0.94\linewidth]{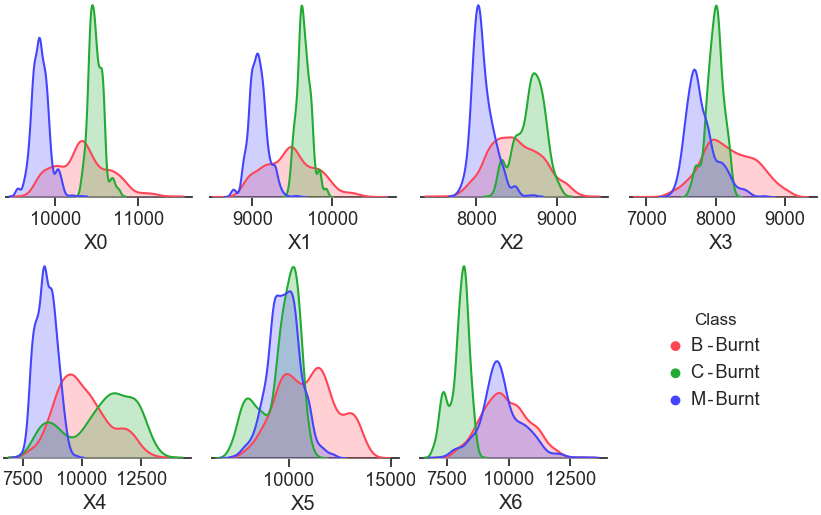}
	\includegraphics[width=0.94\linewidth]{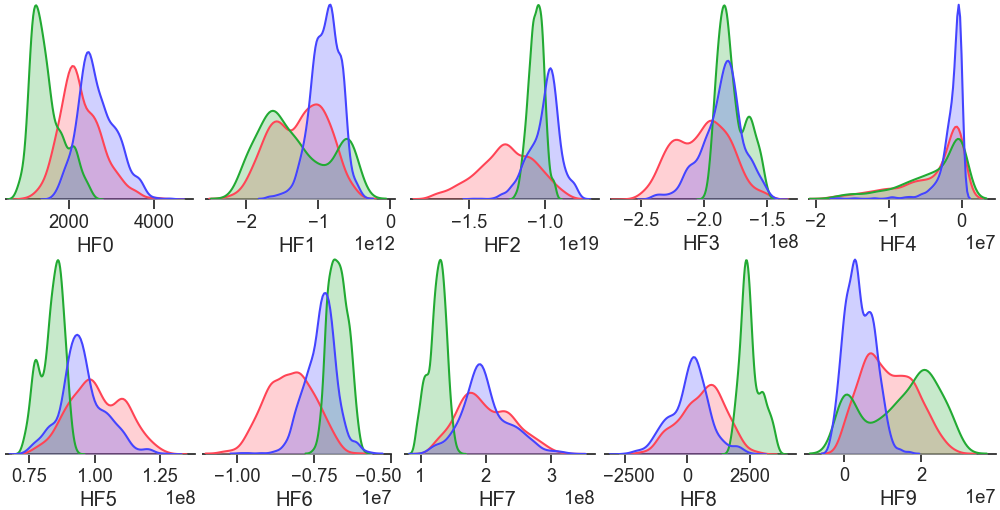} 
	\caption{Distribution of feature values of the burnt class in the original datasets (top half) and in the hyper-datasets (bottom half).}

	\label{BCM_burnt_dist}
\vspace{-15pt}
\end{figure}

Although we are not able to visualize a feature space in $\mathbb{R}^7$, we can still gather important information by simply observing the location and dispersion of the values of the different features. Figure~\ref{M_dist} shows the distribution of the burnt and non-burnt samples of the M dataset. These plots reveal that in 4 out of 7 features, the burnt class is less disperse and is located within the limits of the non-burnt class. The plots of the other datasets (not shown) have similar characteristics. Therefore, it is expected that the ML methods focus their learning in recognizing the burnt class (the positive cases), classifying everything else as non-burnt (the negative cases). This is problematic when training models on one dataset and applying them on a different dataset. Figure~\ref{BCM_burnt_dist} (top half), showing the distribution of burnt samples of the three original datasets, illustrates this problem very well. We can observe that the burnt samples of the C and M datasets almost do not overlap in 4 out of 7 features. The effect of this separation is that the burnt samples of one dataset are likely to be classified as non-burnt by a model trained on the other dataset. This is a plausible explanation for the incompatibility between C and M reported in Sect.~\ref{generalization_outside}. It may also justify why the largest gains in using a mixed dataset happens precisely when training in CM (and testing in B).

Figure~\ref{BCM_burnt_dist} (top half) also reveals that dataset B has the largest dispersion of the burnt class, ``containing'' the C and M burnt values almost entirely, in all features. This tendency is less marked but still visible in the hyper-datasets (same figure, bottom half). Higher dispersion of the burnt class suggests that a higher diversity of burnt types is available for learning, which we expect to result in more generalist models. This may explain why, among the three countries, the specialized models achieving the highest accuracy when tested on other datasets are precisely the ones trained on B and hyper-B. Although these datasets are apparently good sources, they may be bad targets, since specialized models may find it hard to generalize well to a large diversity of burnt types. Indeed, the specialized models tested in B and hyper-B had some of the worst results.

Following the previous logic, one reason for datasets C and hyper-C being bad sources may be the lowest dispersion of their features and hyper-features. 
Mixed datasets increase the dispersion of the burnt classes, and the effect is that the models trained on them can generalize better on new data.

\subsection{Visualizing Class Separability}
\label{class_separability}
In Fig.~\ref{M_dist} we have seen that, on a per-feature basis, the burnt class is less disperse and mostly located inside the limits of the non-burnt class. In Fig.~\ref{BCM_burnt_dist} we have also seen that the burnt classes of the three datasets do not match each other, but the evolved hyper-features seem to reduce this problem (e.g., in hyper-feature HF4 there is very good matching). Taking advantage of the high diversity of models that M3GP can produce in different runs, we have looked among the 30 models produced for each experiment and found some that reduce the dimensionality of the original feature space to only three hyper-features, thus allowing its visualization. We have selected two of these models for illustration purposes. Both have test results above the median in all test cases when compared to the other models from the same set of runs. We intend to visualize the separation of classes inside and outside the training images.

\begin{figure}[]
	\centering
	\includegraphics[width=0.92\linewidth, frame]{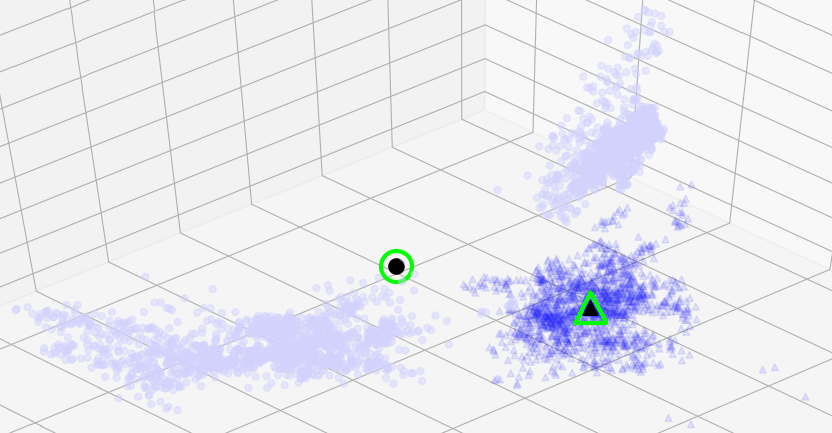}
	\vspace{-8pt}
	\caption{3D Scatter plot of the application of a model trained in the dataset M, on its own test dataset. Burnt and non-burnt samples are drawn in dark and light colors, and triangles and circles represent their centroids, respectively.} 
	\label{M25M_semValores}
\bigskip
	\includegraphics[width=0.92\linewidth, frame]{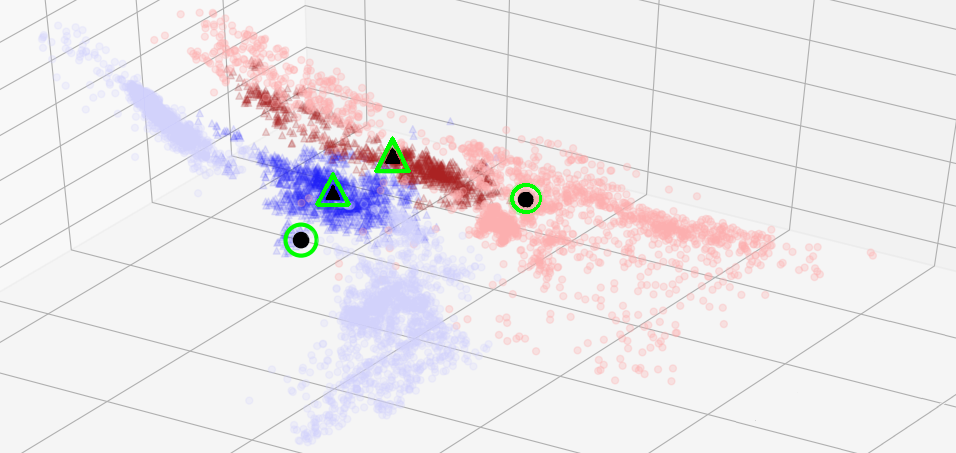}
	\vspace{-8pt}	
	\caption{Application of a model trained in the M dataset in its training set~(blue) and in the C dataset (red). Burnt and non-burnt samples are drawn in dark and light colors, and triangles and circles represent their centroids, respectively.}
	\label{M25C_semValores}
\bigskip
	\includegraphics[width=0.92\linewidth, frame]{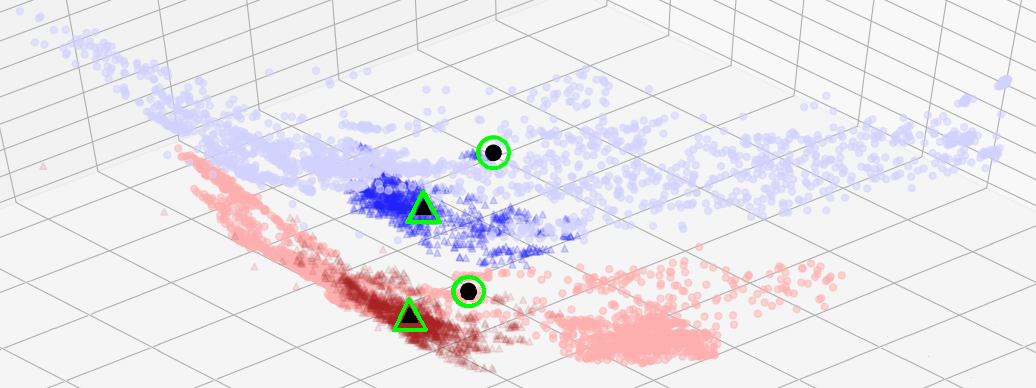}
	\vspace{-8pt}
	\caption{Application of a model trained in the C dataset in its training set~(blue) and in the M dataset (red). Burnt and non-burnt samples are drawn in dark and light colors, and triangles and circles represent their centroids, respectively.} 
	
	\label{C11M_semValores}
\end{figure}

The first model was obtained when training on the M dataset. In Fig.~\ref{M25M_semValores} we see the result of applying this model on its own test set, with the burnt and non-burnt classes represented in dark and light blue, respectively. There seem to be two clusters for the non-burnt class, an observation that is compatible with the two non-burnt peaks visible in most features of Fig.~\ref{M_dist}. However, the MD classifier, which is the fitness function of M3GP, assumes only one cluster per class. The centroids of each cluster are also represented, with a triangle for burnt and a circle for non-burnt.

In Fig.~\ref{M25C_semValores}, we see the result of applying this model on the C dataset. The M training data is represented in blue, the C dataset in red, dark and light colours represent burnt and non-burnt, respectively, and respective centroids as triangles and circles. What we observe is that, although the dispersion of values is similar to the one observed in Fig.~\ref{M25M_semValores}, there is an offset between the class centroids of the M and C datasets. Since the class predictions are based on the distances to the M centroids and the burnt clusters are very compact, this offset causes many burnt samples in C to be misclassified as non-burnt. From the confusion matrix in Table~\ref{confusion} (left matrix), we calculate that the accuracy of this model on C is 80.8\%, and it correctly classifies only half of the burnt samples.

The second model was obtained when training on the C dataset. In Fig.~\ref{C11M_semValores}, we see the result of applying it on the M dataset. Like before, training data is represented in blue, testing data in red; dark and light colours for burnt and non-burnt; triangles and circles for the centroids of burnt and non-burnt. Looking again at Table~\ref{confusion} (right matrix), we calculate that the accuracy of this model on M is 57.6\%, and it correctly classifies only 17.2\% of the burnt samples.

\begin{table}[]
	\centering
	\caption{Confusion matrices of testing in C a model trained in M (left) and testing in M a model trained in C (right).}
	\label{confusion}
    \begin{tabular}{cc}
        \begin{tabular}{l|ccc}
            M $\Rightarrow$ C & $\neg$ Burnt	& Burnt \\%& Total	\\
        	\hline 
            $\neg$ Burnt   & 1858 & 114 \\%& 1972\\
            Burnt       & 433 & 444 \\% & 877\\
            \hline
            Predictions & 2291 & 558
	    \end{tabular}
	    &
        \begin{tabular}{l|ccc}
            C $\Rightarrow$ M & $\neg$ Burnt	& Burnt \\%& Total	\\
	        \hline 
            $\neg$ Burnt   & 1964 & 345\\% & 2309\\
            Burnt       & 1302 & 271  \\%& 1573\\
            \hline
            Predictions & 3266 & 616
	    \end{tabular}
    \end{tabular}
    \vspace{-12pt}
\end{table}

\subsection{Transferring Hyper-features}
%\label{thf}

The previous discussion on class separability revealed that centroid offsets are an obstacle to good generalization on new data. Since the hyper-features were evolved only taking into consideration pure datasets, they are likely the cause of this offset. We have already shown that training in mixed datasets yields better results on a third dataset. However, it is interesting to check what happens in a transfer learning scenario where the hyper-features are evolved on the source dataset, and then transferred and used for further learning on the target dataset (without any additional feature evolution), as described in~\cite{transfer}. For the MD classifier, the only additional learning after adopting the hyper-features is calculating the centroids and covariance matrices of the target dataset on the new hyper-feature space. This means, for the model behind Fig.~\ref{M25C_semValores}, to use the centroids and covariance matrices of C (the target), instead of M (the source). Needless to say, this eliminates the problem of the centroids offset, and the accuracy of this model increases from 80.8\% to 93.06\%. For the model behind Fig.~\ref{C11M_semValores}, using the centroids and covariance matrices of M (the target), instead of C (the source), increases the accuracy from 57.6\% to 97.3\%.

Our goal of providing improved remote sensing models did not initially contemplate the possibility of further learning on the target, since we want to avoid the need to have labelled data on the target. However, in the case of MD, the additional learning requires minimal effort, and although it requires some labelled data, it does not have to be in large quantity, but just enough to calculate centroids and covariance matrices.
Although transfer learning was not in our initial plans, it may be a path we will have to take in pursuit of the ideal classification models. In this work, we still did not perform any proper transfer learning tests (besides the small exercise at the beginning of this section), since the hyper-features evolved were obtained on the mix of three pure datasets, and never transferred for further learning on a fourth dataset.

\subsection{Tailoring Hyper-features}

Regarding the 10 hyper-features evolved for this work, they were tailored to the MD classifier because M3GP uses MD as the fitness function. As expected, they have proven to be more useful to MD than to the other methods (see Table~\ref{pvalues}). Nevertheless, even the method that found them less useful (DT) benefits from using them. The model achieving the highest overall accuracy on the test samples of all three datasets, weighted by the number of samples from each dataset, was XGB using the hyper-features, with 98.05\% (followed closely by XGB using the original features, with 97.7\%).

However, the hyper-features do not have to be tailored to MD. In fact, by replacing the fitness function of M3GP with any other classifier, we can choose for which method the features will be evolved. We expect DT to benefit much more from features that evolved specifically to be used by DT, and we hope that the features evolved for XGB will be able to improve the current results even more.

%% file: Tex/L_conclusions_futurework.tex
\section{Conclusions and Future Work}
\vspace{-3pt}
\par
Intending to improve remote sensing models for classifying burnt areas, this work had two objectives. The first one was to understand the relationship between feature spaces and the predictive ability of the models, to explain the differences in learning and generalization when training and testing in different datasets. We have studied the effects of training a model on data from multiple images with different radiometric characteristics, rather than from a single image. We have found that training on datasets built from more images provides models that generalize better, and that some datasets are better for training while others are better for testing. Visualizing some characteristics of the feature space created by each image dataset, we have intuitively explained the differences in the predictive ability of the evolved models. The second objective was to evolve hyper-features from the original ones, and compare the performance of different methods when using the evolved versus the original features. We have used M3GP both as a baseline and for inducing the hyper-features, and other methods including Decision Trees, Random Forests, and XGBoost, to assess the performance of the hyper-features. They were found to be beneficial for all the methods, and the best model was achieved with XGBoost, even if the hyper-features were tailored to the Mahalanobis Distance classifier used by M3GP.
\par
As future work, we want to explore and visualize other characteristics of the feature spaces that may allow us to predict difficulties in learning and generalization, including in transfer learning scenarios. We also want to improve the generation and selection of hyper-features. On the one hand, we want to impose restrictions on the evolved hyper-features to make them behave more like indices; on the other hand, we want to evolve hyper-features tailored to other methods other than the Mahalanobis Distance classifier. When selecting a set of evolved hyper-features, we want to test other criteria besides the effect they have on fitness, for example, the amount of correlation that exists between each other. Finally, we will extend this work to multiclass classification problems, and we will also consider its applicability to regression problems in the RS domain.

%% file: Tex/O_acknowledgment.tex
\vspace{-4pt}
\section*{Acknowledgment}
\vspace{-3pt}
This work was partially supported by FCT through funding of LASIGE Research Unit (UIDB/00408/2020); projects PTDC/CCI-INF/29168/2017, PTDC/CTA-AMB/30056/2017, PTDC/CCI-CIF/29877/2017, PTDC/ASP-PLA/28726/2017, DSAIPA/DS/0022/2018; PhD Grant SFRH/BD/143972/2019.
\vspace{-4pt}